\documentclass{article}

\usepackage[preprint]{neurips_2025}

\usepackage[utf8]{inputenc} % allow utf-8 input
\usepackage[T1]{fontenc}    % use 8-bit T1 fonts
\usepackage{hyperref}       % hyperlinks
\usepackage{url}            % simple URL typesetting
\usepackage{booktabs}       % professional-quality tables
\usepackage{amsfonts}       % blackboard math symbols
\usepackage{nicefrac}       % compact symbols for 1/2, etc.
\usepackage{microtype}      % microtypography
\usepackage{xcolor}         % colors
\usepackage[nolist,nohyperlinks]{acronym}
\usepackage{mathtools}

\usepackage{microtype}
\usepackage{graphicx}
\usepackage{subfigure}
\usepackage{booktabs} % for professional tables
\usepackage{subcaption}
\usepackage{multicol}
\usepackage[newfloat]{minted}
\usepackage{caption}
\usepackage{algorithm}
\usepackage{algpseudocode}

\algnewcommand{\algorithmicforeach}{\textbf{for each}}
\algdef{SE}[FOR]{ForEach}{EndForEach}[1]
  {\algorithmicforeach\ #1\ \algorithmicdo}% \ForEach{#1}
  {\algorithmicend\ \algorithmicforeach}% \EndForEach

\newenvironment{code}{\captionsetup{type=listing}}{}
\SetupFloatingEnvironment{listing}{name=Source Code}

\title{Hyperbolic Optimization}

\author{%
  Yanke Wang\thanks{This work was funded by InnoHK-HKCRC. The code is available at \url{https://github.com/Seven-year-promise/hyperbolic_diffusion.git}.} \\
  HKCRC \\
  The Hong Kong University of Science and Technology \\ 
  and \\
  BHM, CPCE\\
  The Hong Kong Polytechnic University (PolyU)\\
  \texttt{yanke.wang@cpce-polyu.edu.hk} \\
  \And
  Kyriakos Flouris \\
  MRC Biostatistics Unit \\
  University of Cambridge \\
  \texttt{kf318@cam.ac.uk} \\
}

\begin{document}

\maketitle

\begin{abstract}
This work explores optimization methods on hyperbolic manifolds. Building on Riemannian optimization principles, we extend the Hyperbolic Stochastic Gradient Descent (a specialization of Riemannian SGD) to a Hyperbolic Adam optimizer. While these methods are particularly relevant for learning on the Poincaré ball, they may also provide benefits in Euclidean and other non-Euclidean settings, as the chosen optimization encourages the learning of Poincaré embeddings. This representation, in turn, accelerates convergence in the early stages of training, when parameters are far from the optimum. As a case study, we train diffusion models using the hyperbolic optimization methods with hyperbolic time-discretization of the Langevin dynamics, and show that they achieve faster convergence on certain datasets without sacrificing generative quality.

% Most of current deep learning models are optimized in Euclidean space, but they may suffer from the non-Euclidean-distributed data/noise structure, especially the hierarchical structure such as tree data/noise learning or graph learning. The data information is distributed unevenly, resulting in the failure of the Euclidean assumption, however, perfectly applicable to hyperbolic space. This paper proposes to refine the model optimization procedure to hyperbolic space, making learning process more efficiently and generating more hyperbolic data-suited models. We select Denoising Diffusion Probabilistic Models (DDPM) as a case study to conduct the image generation task. The objective function, optimizing procedure, and timestep sampler in hyperbolic space are explored with verification results on public datasets. The hyperbolic diffusion models not only achieve faster convergence but good good image generation quality. This work is aimed at providing a guidance for current researchers to study generative models under different mathematical assumptions.
\end{abstract}

\begin{acronym}
\acrodef{ddpm}[DDPM]{Denoising Diffusion Probabilistic Models}
\acrodef{sgd}[SGD]{Stochastic Gradient Descent}
\acrodef{fid}[FID]{Fréchet Inception Distance}
\acrodef{llm}[LLMs]{Large Language Models}
\acrodef{cnn}[CNNs]{Convolutional Neural Networks}
\acrodef{gnn}[GNNs]{Graph Neural Networks}
\end{acronym}

\section{Introduction}
Recently, the training of intelligent models, e.g., vision models, \ac{llm}, and generative models, still rely on the gradient descending or ascending manner thanks to their computational efficiency \cite{wang2025survey}, with the aim lay on the convergence of the objective function. The research on the optimization procedure has been focused on the tuning hyperparameter \cite{liao2022empirical,sambharya2025learning}, adding regularization term \cite{foret2020sharpness}, momentum optimization \cite{li2024context}. Most optimization schemes are built on the assumption of a Euclidean parameter space. While this is reasonable given that the underlying geometry of the optimization manifold is generally unknown, exploring alternative spaces—such as the Poincaré ball, can be advantageous. This is motivated both by the structure of certain data (e.g., hierarchical or tree-like data naturally suited to hyperbolic geometry) and by the properties of hyperbolic spaces themselves, such as their ability to induce accelerated optimization trajectories.
%the Euclidean deep learning models have shown increasing uncertainty of tackling the real-world problems, since the optimal models obtained by Euclidean optimization explicit unsatisfactory robustness, limited enhanceable accuracy, unproven fit into real-world application. This challenge has gained attention from rising researchers. Thus, considering designing and optimizing intelligent models in an alternative mathematical space is proven indispensable.

Previous work has adapted neural architectures to operate in hyperbolic space, including hyperbolic CNNs and GNNs \cite{mettes2024hyperbolic}. As many studies consider the operation of feature encoding in the tangent space, \cite{chen2021fully} achieves a fully hyperbolic neural networks. These models often yield improved representations by learning embeddings that better reflect the underlying geometry of the data \cite{flouris2023canonical, nickel2017poincare, woo2025hypevpr, mossi_matrix_2025}. Neural networks in hyperbolic space have been applied in varying domains, e.g., lower-dimensional embedding feature of magnetoencephalography (MEG)-generated functional brain connectivity graphs \cite{ramirez2025fully}.

Optimization in hyperbolic space has primarily followed two directions: defining loss functions directly on hyperbolic manifolds and developing Riemannian extensions of standard optimizers. In particular, for hierarchical data such as tree-structured tasks, hyperbolic models trained with manifold-aware loss functions align naturally with the geometry and achieve superior performance \cite{ayoughi2025continual}.

% Many pioneers conduct modification by designing the networks in the hyperbolic space, e.g., hyperbolic \ac{cnn} and \ac{gnn} \cite{mettes2024hyperbolic}. The hyperbolic networks generate better data embedding (feature representation) \cite{nickel2017poincare, woo2025hypevpr}. The optimization in hyperbolic space has two manners, i.e., designs of hyperbolic objective functions and optimizers. Especially, the instance classification of hierarchical data, e.g., tree-like structures, a hyperbolic model optimized by a hyperbolic loss function naturally fits this problem to achieve better performance \cite{ayoughi2025continual}.

% So far, the research focused on the optimizers in the hyperbolic space is still insufficient. With the inspiration by the improvement of classification models and data embedding on hyperbolic space, we consider doing the model optimization in hyperbolic space. We summarize our contribution as
% \begin{itemize}
%     \item Designs of hyperbolic \ac{sgd} and AdamW;
%     \item Optimization of \ac{ddpm} via the proposed two optimizers;
%     \item Experiments of image generation by using the optimized models to verify the improvement of the proposed two optimizers.
% \end{itemize}

Research on optimization methods in hyperbolic space remains limited. \cite{pachica2024improved} applies the Hyperbolic Tangent (tanh) functionto the warn-up phase of the training. \cite{ma2018quasi} proposes quasi-hyperbolic momentum (QHM) for the usage in the gradient-descending manner and improves the trainign efficiency.  Focused on large-scale multiobjective optimization, \cite{li2024hyperbolic} designs a hyperbolic preselection operator to better the search efficiency in the mentioned optimization process. Motivated by the improvements observed in classification and representation learning with hyperbolic geometry, we investigate model optimization directly on the Poincaré ball. In this work, our primary contribution is the development of a Riemannian extension of \textit{AdamW on the Poincaré ball}, building on existing work on Riemannian SGD \cite{nickel2017poincare}. To demonstrate its effectiveness, we apply these optimizers to train DDPMs under hyperbolic dynamics and evaluate the resulting models on image generation tasks, showing that they achieve faster convergence and improved sample quality.

\section{Preliminaries}
\label{sec:pre}
Minimization of the loss function is achieved via gradient descent. Assume that the objective function is $f(\theta)$ with parameters $\theta$ to be trained. The update of $\theta$ is given by the vanilla \ac{sgd} \cite{sutskever2013importance} as
\begin{equation}
    \begin{aligned}
        g_t = \nabla f_t(\theta_{t-1}) \\
        \theta_t = \theta_{t-1} - \gamma g_t,
    \end{aligned}
\label{equ:sgd}
\end{equation}
where $\gamma > 0$ is the learning rate, and $\nabla f_t(\theta_{t-1})$ is the gradient at $\theta_{t-1}$.

Additionally, the vanilla Adam improves the performance of \ac{sgd} \cite{kingma2014adam}. The weight decay $\lambda$ and momentum $m_t$ are added to the algorithm,
\begin{equation}
    \begin{aligned}
        \theta_t = \theta_{t-1} - \gamma \lambda \theta_{t-1}, \\
        m_t = \beta_1 m_{t-1} + (1-\beta_1)g_t, \\
        v_t = \beta_2 v_{t-1} + (1-\beta_2) g_t^2, \\
        \hat{m}_t = m_t / (1-\beta_1^t), \\
        \hat{v}_t = v_t / (1-\beta_2^t), \\
        \theta_t = \theta_t - \gamma \hat{m}_t / (\sqrt{\hat{v}_t} + \xi).
    \end{aligned}
\label{equ:adamw}
\end{equation}

In the AdamW variant \cite{loshchilov2017decoupled}, weight decay does not accumulate in the momentum or variance.

In this work, we consider optimization on a Poincaré ball. Given a Poincaré manifold $\mathfrak{B}^d$, the distance between points $\boldsymbol{u}, \boldsymbol{v} \in \mathfrak{B}^d$ is computed as \cite{nickel2017poincare}
\begin{equation}
    d(\boldsymbol{u}, \boldsymbol{v}) = \operatorname{arcosh} \left( 1 + 2 \frac{||\boldsymbol{u} - \boldsymbol{v}||^2}{(1 - ||\boldsymbol{u}||^2)(1 - ||\boldsymbol{v}||^2)}\right),
\end{equation}
where $||\cdot||$ denotes the Euclidean $L_2$-norm. Given the parameters $\theta_t$ to be trained, the optimization update of $\theta_t$ is formulated as
\begin{equation}
    \theta_{t+1} \longleftarrow \text{Proj} \left( \theta_t - \eta_t \frac{(1-||\theta_t||^2)^2}{4} \nabla E\right),
\end{equation}
with $\nabla E$ as the Euclidean gradient and the projection defined as
\begin{equation}
      \text{Proj} (\theta) = \begin{cases}
    \theta / ||\theta|| - \epsilon & \text{if $||\theta|| \geq 0$}, \\
    \theta & \text{otherwise}.
  \end{cases}
\end{equation}

Poincaré embeddings are effective at learning hierarchical representations, which can potentially improve the optimization process during model training.

\section{Hyperbolic Optimization}

\begin{figure}
    \centering
    \includegraphics[width=5.5in]{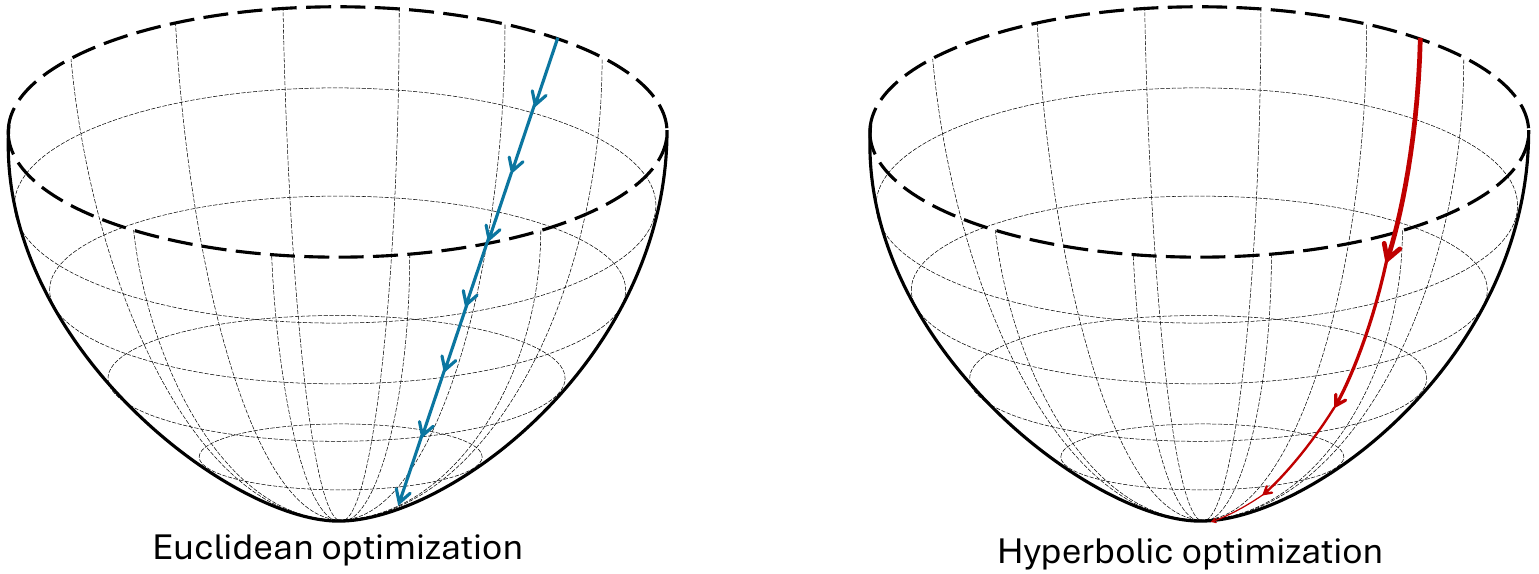}
    \caption{The Euclidean and hyperbolic optimizations.}
    \label{fig:hyperbolic_optimization}
\end{figure}

Given the ability of the Poincaré ball to structure hyperbolic data, optimization can be performed on the Poincaré ball using an AdamW optimizer, similar to Riemannian SGD. Such optimization encourages embedding of the data along a hyperbolic trajectory \cite{nickel2017poincare}. Fig. \ref{fig:hyperbolic_optimization} illustrates the possible paths for finding optimal parameters in Euclidean versus hyperbolic spaces. When the parameter space does not follow a Euclidean geometry, the training process can converge to suboptimal points or fail to converge effectively. Additionally, due to the exponential metric of hyperbolic space, optimization along the hyperbolic trajectory can result in larger effective gradient steps when parameters are far from the optimum, potentially accelerating convergence in the early stages of training.

To implement and verify this idea, we select \ac{ddpm} as a case study. \ac{ddpm} proposed to generate images by denoising the noises from the standard normal distribution step by step \cite{ho2020denoising}. The models naturally needs to a noise scheduler, timestamp sampler for noise addition in each step, and optimizers for the training procedure, this work aims to analyze how to apply the hyperbolic embedding concept to \ac{ddpm} for optimization. \ac{ddpm} works with two processes, e.g., the forward diffusion process and backward denoising process. The noise is added gradually in the forward process, and the timestep sampler follows a uniform (linear) distribution \cite{ho2020denoising}, that is to say, the diffusion process follows an Euclidean as
\begin{equation}
    t \sim \mathcal{U}\{1, T\},
\label{equ:t_linear}
\end{equation}
where $T$ indicates the total timesteps of the diffusion process. To do the timestep sampling on a hyperbolic space, we consider a unit hyperbola and select the non-negative branch of the hyperbola and let the steps ($s$) and timesteps ($t$) have the following relation,
\begin{equation}
    t = \sqrt{s^2 - 1}, \quad s \in \left[1, \sqrt{T^2+1} \right].
\label{equ:t_unit_hyperbola}
\end{equation}

The two main components of optimizing a model are the objective function and parameter update. Given the \ac{ddpm} with parameters $\boldsymbol{\epsilon}_{\theta}$, the initial loss function is by computing the MSE loss as 
\begin{equation}
    \mathcal{L} (\theta) = ||\boldsymbol{\epsilon} -  \boldsymbol{\epsilon}_{\theta} ||^2, \quad \boldsymbol{\epsilon} \sim \mathcal{N}(\boldsymbol{0}, \boldsymbol{\text{I}}).
\label{equ:loss}
\end{equation}

According to the hyperbolic distance defined in Section \ref{sec:pre}, we propose to rewrite the loss function of \ac{ddpm} on Poincaré ball as 
\begin{equation}
    \mathcal{L}_\mathfrak{B} (\theta) = \text{arcosh} \left( 1 + 2 \frac{||\boldsymbol{\epsilon} -  \boldsymbol{\epsilon}_{\theta} ||^2}{(1 - ||\boldsymbol{\epsilon}||^2)(1 - ||\boldsymbol{\epsilon}_{\theta}||^2)}\right).
\label{equ:loss_poincare}
\end{equation}

Following the gradient descending manner, we need to compute the gradient on the hyperbolic space given the Euclidean gradient $g_t$ via the following transformation and projection,
\begin{equation}
    \mathfrak{g}_t = \frac{(1-||\theta_{t-1}||^2)^2}{4} g_t ,
\label{equ:hyper_gt}
\end{equation}
\begin{equation}
    \theta_t = \text{Proj} \left( \theta_{t-1} - \gamma \mathfrak{g}_t \right) .
\label{equ:projection}
\end{equation}

We chose to explore two optimizers, i.e. Hyperbolic \ac{sgd} and Hyperbolic AdamW, with the hyperbolic gradient. The update of parameters via hyperbolic \ac{sgd} is completed by Algorithm \ref{alg:hyper_sgd} and hyperbolic AdamW by Algorithm \ref{alg:hyper_adamw}, with corresponding source codes in Appendix \ref{appen:code}.

% Moreover, the update of parameters via hyperbolic AdamW follows
% \begin{equation}
%     \begin{aligned}
%         \theta_t = \theta_{t-1} - \gamma \lambda \theta_{t-1} \\
%         m_t = \beta_1 m_{t-1} + (1-\beta_1) \mathfrak{g}_t \\
%         v_t = \beta_2 v_{t-1} + (1-\beta_2) \mathfrak{g}_t^2 \\
%         \hat{m_t} = m_t / (1-\beta_1^t) \\
%         \hat{v_t} = v_t / (1-\beta_2^t) \\
%         \theta_t = \text{Proj}\left(\theta_t - \gamma \hat{m_t} / (\sqrt{\hat{v_t}} + \xi)\right)
%     \end{aligned}
% \label{equ:hyper_adamw}
% \end{equation}.

\begin{algorithm}
\caption{Hyperbolic \ac{sgd}}\label{alg:hyper_sgd}
\begin{algorithmic}
\Require{$\gamma$ \hspace{0.05em} \text{(lr)}, \quad $\theta_{t-1}$ \hspace{0.05em} \text{(parameters)}.} \;

\For{$t=1$ \hspace{0.05em} \textbf{to} \hspace{0.05em} ...}
\State $g_t \gets \nabla \mathcal{L}(\theta_{t-1})$

\State $\mathfrak{g}_t \gets \frac{(1-||\theta_{t-1}||^2)^2}{4} g_t$  \algorithmiccomment{Refer to Eq. \ref{equ:hyper_gt}.}

\State $\theta_t \gets \text{Proj} \left( \theta_{t-1} - \gamma \mathfrak{g}_t \right)$  \algorithmiccomment{Refer to Eq. \ref{equ:projection}.} 
\EndFor
\Ensure $\theta_t$
\end{algorithmic}
\end{algorithm}

\begin{algorithm}
\caption{Hyperbolic AdamW}\label{alg:hyper_adamw}
\begin{algorithmic}
\Require{$\gamma$ \hspace{0.05em} \text{(lr)}, \quad $\theta_{t-1}$ \hspace{0.05em} \text{(parameters)},  \quad $\beta_1$, $\beta_2$ \hspace{0.05em} \text{(betas)},  \quad $\lambda$ \hspace{0.05em} \text{(weight decay)},  \quad poincare\_loss.} \;

\If{poincare\_loss}
\State $g_t \gets \nabla \mathcal{L}(\theta_{t-1})$ \algorithmiccomment{Refer to Eq. \ref{equ:loss}.}
\Else
\State $g_t \gets \nabla \mathcal{L}_\mathfrak{B}(\theta_{t-1})$ \algorithmiccomment{Refer to Eq. \ref{equ:loss_poincare}.}
\EndIf

\For{$t=1$ \hspace{0.05em} \textbf{to} \hspace{0.05em} ...}

\State $\mathfrak{g}_t \gets \frac{(1-||\theta_{t-1}||^2)^2}{4} g_t$  \algorithmiccomment{Refer to Eq. \ref{equ:hyper_gt}.}

\State $\theta_t \gets \theta_{t-1} - \gamma \lambda \theta_{t-1}$

\State $m_t \gets \beta_1 m_{t-1} + (1-\beta_1) \mathfrak{g}_t$

\State $v_t \gets \beta_2 v_{t-1} + (1-\beta_2) \mathfrak{g}_t^2$ 

\State $\hat{m_t} \gets m_t / (1-\beta_1^t)$ 

\State $\hat{v_t} \gets v_t / (1-\beta_2^t)$

\State $\theta_t \gets \text{Proj}\left(\theta_t - \gamma \hat{m_t} / (\sqrt{\hat{v_t}} + \xi)\right)$ \algorithmiccomment{Refer to Eq. \ref{equ:projection}.} 
\EndFor
\Ensure $\theta_t$
\end{algorithmic}
\end{algorithm}
A Python implementation is provided in Appendix \ref{appen:code}, which can be used to straightforwardly adapt the original AdamW to Hyperbolic AdamW.

% \textcolor{red}{Highlight and how people cna use these equations, hive the codes in the paper.}

\section{Experiment and Verification}

We choose the image generation task to verify the optimized model on generation ability. The dataset is a sub-set of butterfly images from Smithsonian Institution \cite{smithsonian_butterflies_subset}, containing 1000 images. We train the model on GeForce RTX 4090D. The \ac{fid} score is computed between the training set and generated images.

The two vanilla optimizers accompany with different optimal hyper-parameters, so we conduct two group of experiments. The vanilla \ac{sgd} and hyperbolic \ac{sgd} (Group 1) are compared by
\begin{itemize}
    \item \ac{sgd} optimizer with a linear timestep sampler (SGD+LinearT),
    \item Hyperbolic \ac{sgd} optimizer with a linear timestep sampler (HyperSGD+LinearT), and
    \item Hyperbolic \ac{sgd} optimizer with a timestep sampler in the function of unit hyperbola (HyperSGD+HyperT),
\end{itemize}
in which the learning rate is set as 0.002, and models are trained for 500 epochs until convergence. Notably, the models begin to converge at 300th epoch, but they are trained for more iterations to guarantee the fairness for comparison. The inference steps are set as 200 steps. Additionally, the vanilla AdamW and hyperbolic AdamW (Group 2) are compared by
\begin{itemize}
    \item AdamW optimizer with a linear timestep sampler (AdamW+LinearT),
    \item Hyperbolic AdamW optimizer with a timestep sampler in the function of unit hyperbola (HyperAdamW+HyperT), and
    \item Hyperbolic AdamW optimizer with a timestep sampler in the function of unit hyperbola and a loss function on Poincaré (HyperAdamW+HyperT+HyperLoss),
\end{itemize}
in which the learning rate is set as 0.0002, and models are trained for 350 epochs until convergence. The inference steps are 200 and 50 steps, respectively.

\subsection{Result and Analysis}
\label{sec:exp_res}
In the experiment Group 1, the comparison of SGD+LinearT, HyperSGD+LinearT, and HyperSGD+HyperT are drawn in Fig. \ref{fig:sgd_fid_epoch_200steps} with smoothed plot and the generated images are visualized in Appendix \ref{appen:gene_sgd}. It is clearly seen that the optimizer in hyperbolic space HyperSGD+HyperT achieves faster convergence and lower \ac{fid} score. However, given the same optimizer, the timestep sampler in a unit hyperbola function (HyperT) helps in accelerating the training, but not in lowering the \ac{fid} score. 

\begin{figure}[h]
  \centering
  \includegraphics[width=4.3in]{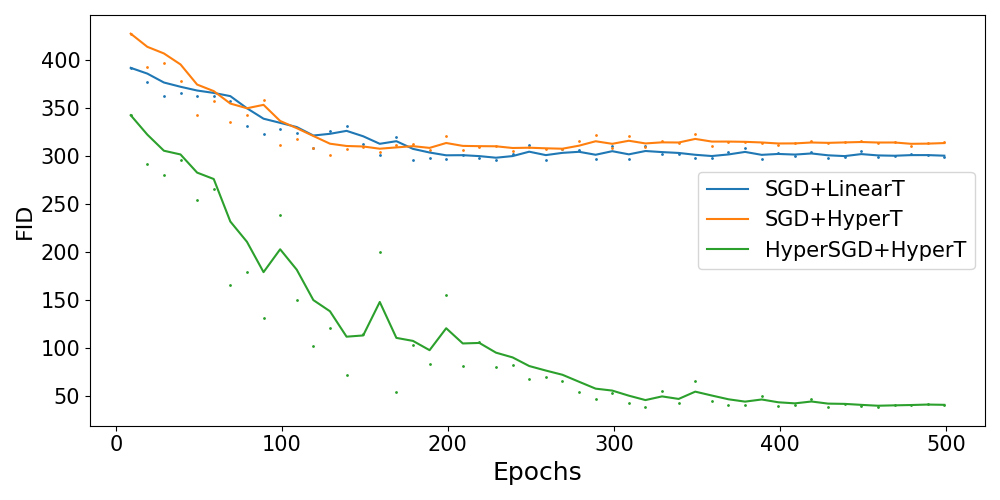}
  \caption{FID Comparison of \ac{sgd} and hyperbolic \ac{sgd} optimizers with 200 inference steps.}
  \label{fig:sgd_fid_epoch_200steps}
\end{figure}

\begin{figure}[h]
  \centering
    \subfigure[FID Comparison with 200 inference steps.]{\includegraphics[width=0.47\textwidth]{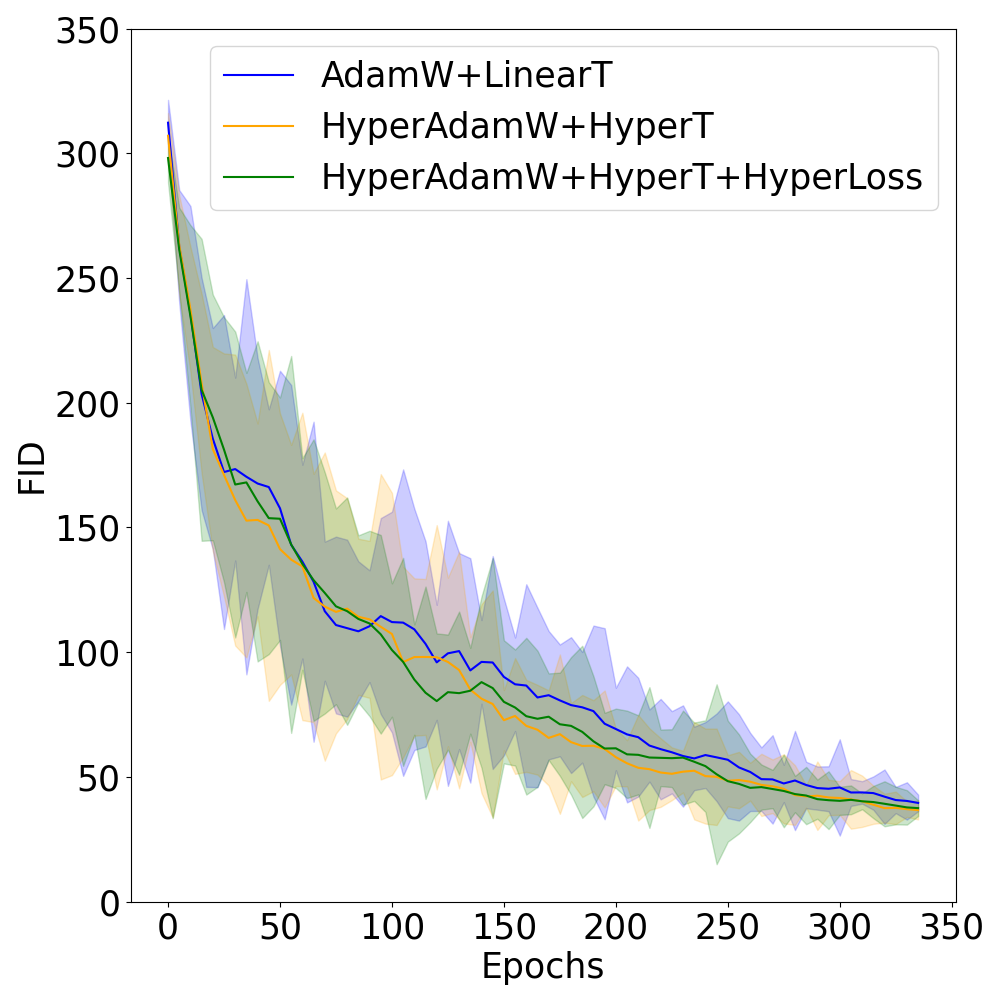}
  \label{fig:comarison_optimizers_fid_200steps}}
  \centering
  \subfigure[FID Comparison 50 inference steps.]{\includegraphics[width=0.47\textwidth]{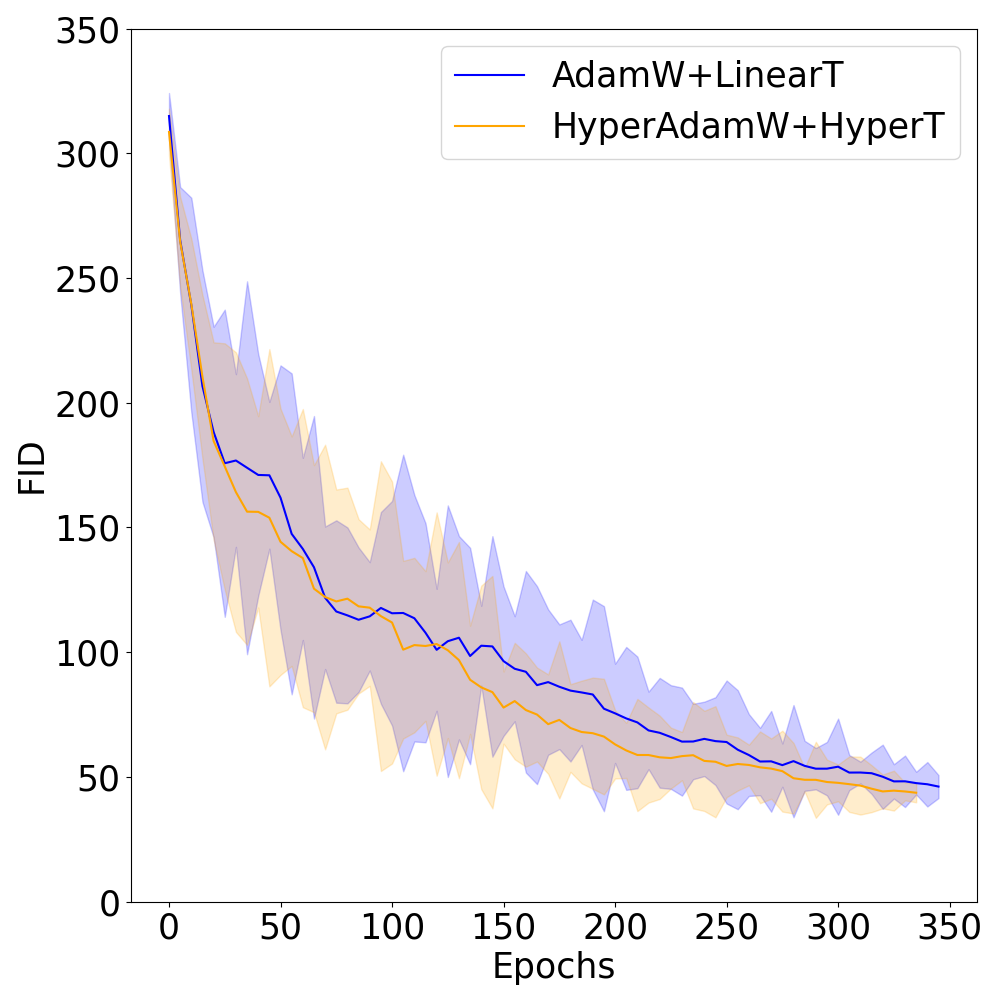}
  \label{fig:comarison_optimizers_fid_50steps}}
  \caption{FID Comparison of AdamW and hyperbolic AdamW optimizers with 200 and 50 inference steps, respectively.}
  \label{fig:comarison_optimizers}
\end{figure}

Furthermore, we consider proving the generalization of the idea on other optimizers with diverse configurations, so the AdamW+LinearT, HyperAdamW+HyperT, and HyperAdamW+HyperT+HyperLoss are also compared as Group 2. The results of \ac{fid} are plotted in Fig. \ref{fig:comarison_optimizers_fid_200steps} with 200 inference steps and Fig. \ref{fig:comarison_optimizers_fid_50steps} with 50 inference steps. The HyperAdamW can also put the convergence in advance but not significantly. In this case, we consider that this improvement may happen by coincidence, so we run the experiments 10 times and visualize the mean plots with the corresponding variances as in Fig. \ref{fig:comarison_optimizers_fid_200steps} and \ref{fig:comarison_optimizers_fid_50steps}. The image generations by one of the experiment are outlined in Appendix \ref{appen:gene_adamw}. The replicates of the experiments verify our findings in Group 1.

\subsection{Limitations}
\label{sec:exp_limit}
Although our implementation and extension of the methods have improvements on the training of models, we only verify this finding on two optimizers and one dataset. Thus, more datasets are expected to be involved in these experiments. Additionally, we select the diffusion model as a case study, but the optimizers are potential to help any other models, e.g., image classification, object detection and segmentation, \ac{llm}, etc.

\section{Conclusion}
\label{sec:conclusion}
Focusing on the convergence of deep learning models, we explore and implement an AdamW-based optimization procedure in hyperbolic space. Using a case study on image generation with diffusion models, we evaluate the optimizers under different conditions. The results indicate that, in certain cases, the hyperbolic optimizers can accelerate the convergence of diffusion models and also improve generation performance.

\bibliographystyle{abbrv}
\bibliography{reference}

\begin{thebibliography}{10}

\bibitem{smithsonian_butterflies_subset}
\url{https://huggingface.co/datasets/huggan/smithsonian_butterflies_subset/}.

\bibitem{ayoughi2025continual}
M.~Ayoughi, M.~G. Atigh, M.~M. Derakhshani, C.~G. Snoek, P.~Mettes, and
  P.~Groth.
\newblock Continual hyperbolic learning of instances and classes.
\newblock {\em arXiv preprint arXiv:2506.10710}, 2025.

\bibitem{chen2021fully}
W.~Chen, X.~Han, Y.~Lin, H.~Zhao, Z.~Liu, P.~Li, M.~Sun, and J.~Zhou.
\newblock Fully hyperbolic neural networks.
\newblock {\em arXiv preprint arXiv:2105.14686}, 2021.

\bibitem{flouris2023canonical}
K.~Flouris et~al.
\newblock Canonical normalizing flows for manifold learning.
\newblock {\em NeurIPS}, 2023.

\bibitem{foret2020sharpness}
P.~Foret, A.~Kleiner, H.~Mobahi, and B.~Neyshabur.
\newblock Sharpness-aware minimization for efficiently improving
  generalization.
\newblock {\em arXiv preprint arXiv:2010.01412}, 2020.

\bibitem{ho2020denoising}
J.~Ho, A.~Jain, and P.~Abbeel.
\newblock Denoising diffusion probabilistic models.
\newblock {\em Advances in neural information processing systems},
  33:6840--6851, 2020.

\bibitem{kingma2014adam}
D.~P. Kingma and J.~Ba.
\newblock Adam: A method for stochastic optimization.
\newblock {\em arXiv preprint arXiv:1412.6980}, 2014.

\bibitem{li2024hyperbolic}
B.~Li, Y.~Yang, W.~Hong, P.~Yang, and A.~Zhou.
\newblock Hyperbolic neural network based preselection for expensive
  multi-objective optimization.
\newblock {\em IEEE Transactions on Evolutionary Computation}, 2024.

\bibitem{li2024context}
D.~Li, Z.~Liu, X.~Hu, Z.~Sun, B.~Hu, and M.~Zhang.
\newblock In-context learning state vector with inner and momentum
  optimization.
\newblock {\em Advances in Neural Information Processing Systems},
  37:7797--7820, 2024.

\bibitem{liao2022empirical}
L.~Liao, H.~Li, W.~Shang, and L.~Ma.
\newblock An empirical study of the impact of hyperparameter tuning and model
  optimization on the performance properties of deep neural networks.
\newblock {\em ACM Transactions on Software Engineering and Methodology
  (TOSEM)}, 31(3):1--40, 2022.

\bibitem{loshchilov2017decoupled}
I.~Loshchilov and F.~Hutter.
\newblock Decoupled weight decay regularization.
\newblock {\em arXiv preprint arXiv:1711.05101}, 2017.

\bibitem{ma2018quasi}
J.~Ma and D.~Yarats.
\newblock Quasi-hyperbolic momentum and adam for deep learning.
\newblock {\em arXiv preprint arXiv:1810.06801}, 2018.

\bibitem{mettes2024hyperbolic}
P.~Mettes, M.~Ghadimi~Atigh, M.~Keller-Ressel, J.~Gu, and S.~Yeung.
\newblock Hyperbolic deep learning in computer vision: A survey.
\newblock {\em International Journal of Computer Vision}, 132(9):3484--3508,
  2024.

\bibitem{mossi_matrix_2025}
A.~Mossi, B.~Žunkovič, and K.~Flouris.
\newblock A matrix product state model for simultaneous classification and
  generation.
\newblock {\em Quantum Machine Intelligence}, 7(1):48, Apr. 2025.

\bibitem{nickel2017poincare}
M.~Nickel and D.~Kiela.
\newblock Poincar{\'e} embeddings for learning hierarchical representations.
\newblock {\em Advances in neural information processing systems}, 30, 2017.

\bibitem{pachica2024improved}
A.~O. Pachica, A.~C. Fajardo, and R.~P. Medina.
\newblock Improved adam optimizer with warm-up strategy and hyperbolic tangent
  function for employee turnover prediction using multilayer perceptron (mlp).
\newblock In {\em 2024 15th International Conference on Information and
  Communication Technology Convergence (ICTC)}, pages 283--288. IEEE, 2024.

\bibitem{ramirez2025fully}
H.~Ramirez, D.~Tabarelli, A.~Brancaccio, P.~Belardinelli, E.~B. Marsh,
  M.~Funke, J.~C. Mosher, F.~Maestu, M.~Xu, and D.~Pantazis.
\newblock Fully hyperbolic neural networks: A novel approach to studying aging
  trajectories.
\newblock {\em IEEE Journal of Biomedical and Health Informatics}, 2025.

\bibitem{sambharya2025learning}
R.~Sambharya, J.~Bok, N.~Matni, and G.~Pappas.
\newblock Learning acceleration algorithms for fast parametric convex
  optimization with certified robustness, 2025.

\bibitem{sutskever2013importance}
I.~Sutskever, J.~Martens, G.~Dahl, and G.~Hinton.
\newblock On the importance of initialization and momentum in deep learning.
\newblock In {\em International conference on machine learning}, pages
  1139--1147. PMLR, 2013.

\bibitem{wang2025survey}
J.~Wang and A.~Choromanska.
\newblock A survey of optimization methods for training dl models: Theoretical
  perspective on convergence and generalization.
\newblock {\em arXiv preprint arXiv:2501.14458}, 2025.

\bibitem{woo2025hypevpr}
S.~Woo, S.~Lee, J.~Jang, and E.~Kim.
\newblock Hypevpr: Exploring hyperbolic space for perspective to
  equirectangular visual place recognition.
\newblock {\em arXiv preprint arXiv:2506.04764}, 2025.

\end{thebibliography}

\appendix
\section{Image Generation of Models Trained by \ac{sgd}}
\label{appen:gene_sgd}

\begin{figure}[h!]
    \centering
    \subfigure[10]{\includegraphics[width=0.11\textwidth]{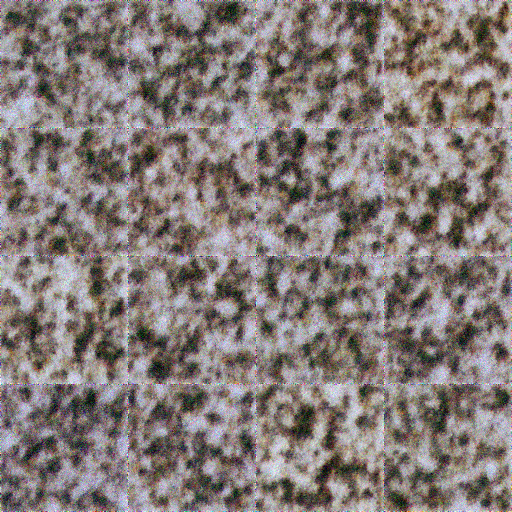}} 
    \subfigure[20]{\includegraphics[width=0.11\textwidth]{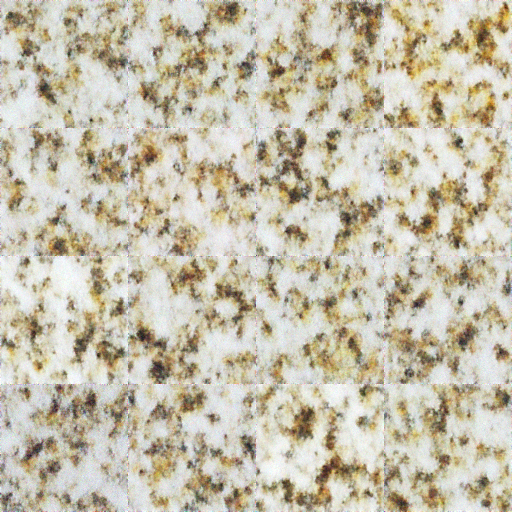}}
    \subfigure[30]{\includegraphics[width=0.11\textwidth]{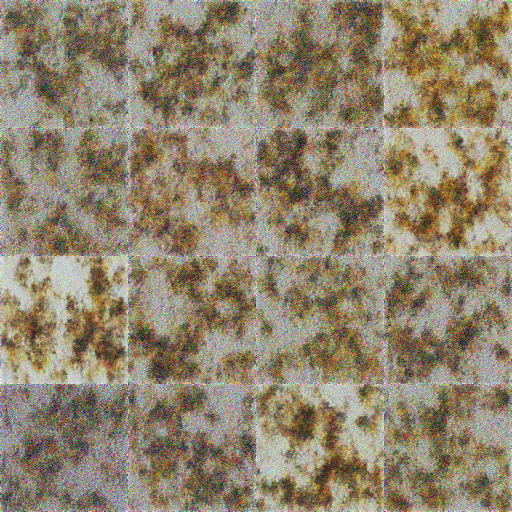}}
    \subfigure[50]{\includegraphics[width=0.11\textwidth]{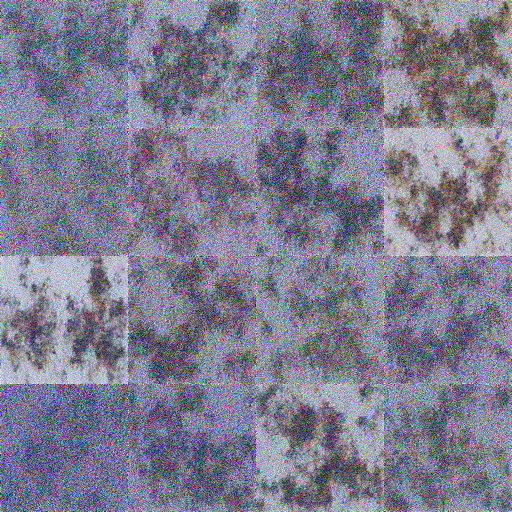}}
    \subfigure[100]{\includegraphics[width=0.11\textwidth]{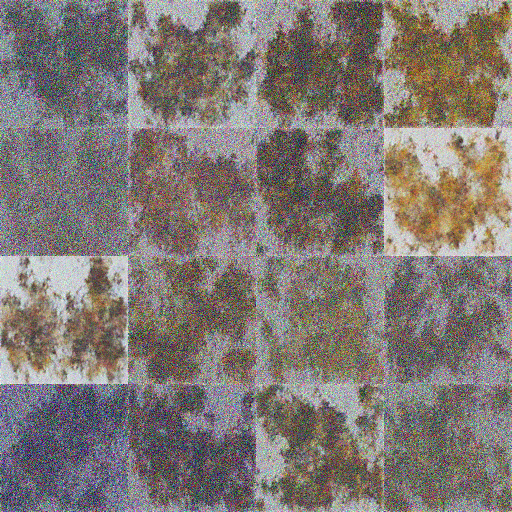}}
    \subfigure[200]{\includegraphics[width=0.11\textwidth]{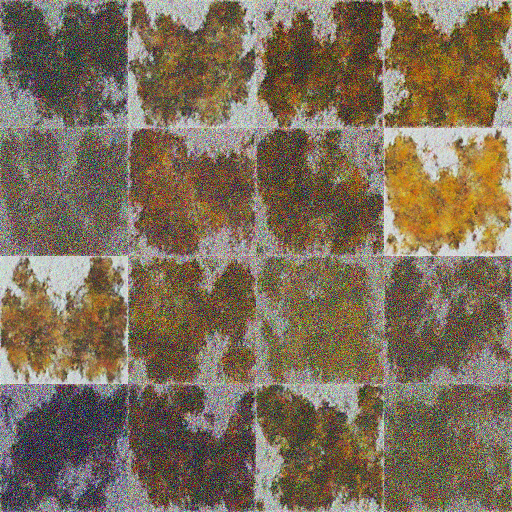}}
    \subfigure[300]{\includegraphics[width=0.11\textwidth]{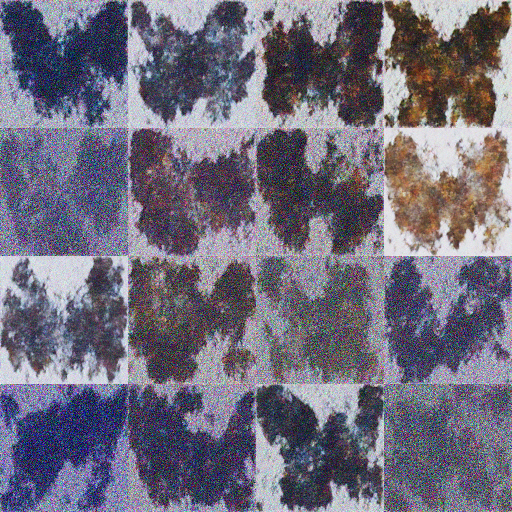}}
    \subfigure[500]{\includegraphics[width=0.11\textwidth]{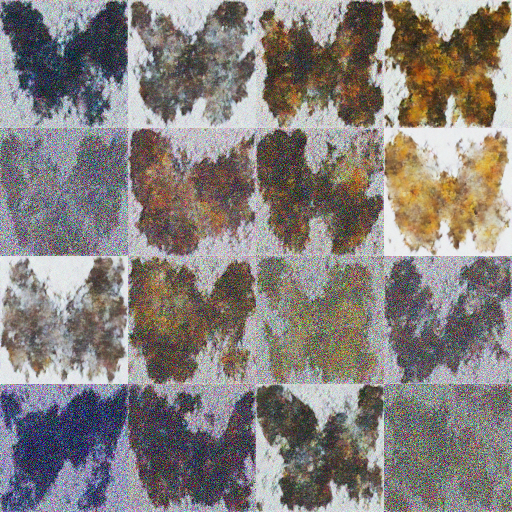}}
    \caption{Training DDPM via SGD (lr=2e-3), increasing epochs (10, 20, ..., 500).}
    \label{fig:ddpm_sgd_Tlinear0.002}
\end{figure}

\begin{figure}[h!]
    \centering
    \subfigure[10]{\includegraphics[width=0.11\textwidth]{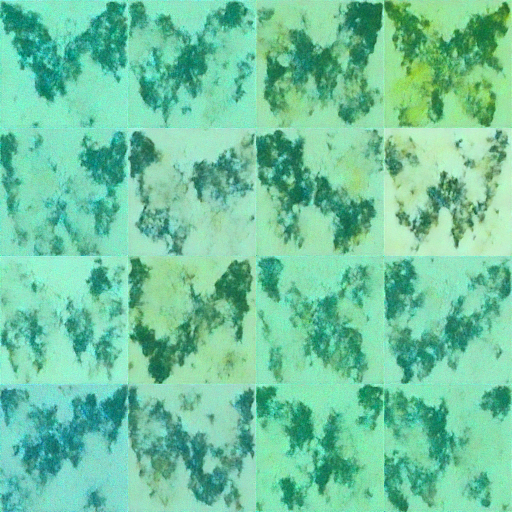}} 
    \subfigure[20]{\includegraphics[width=0.11\textwidth]{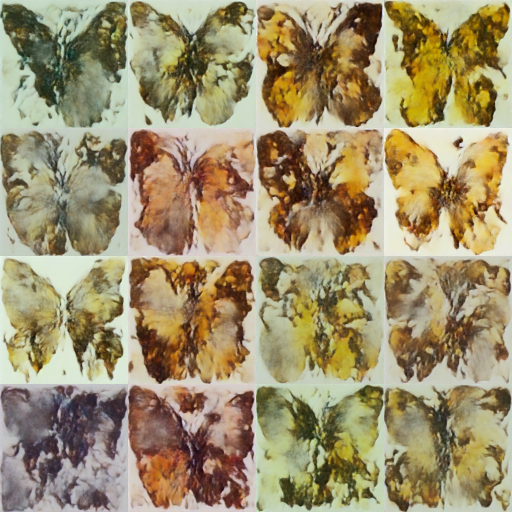}}
    \subfigure[30]{\includegraphics[width=0.11\textwidth]{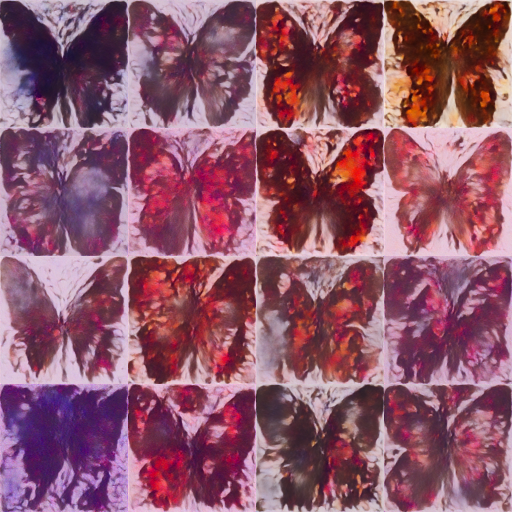}}
    \subfigure[50]{\includegraphics[width=0.11\textwidth]{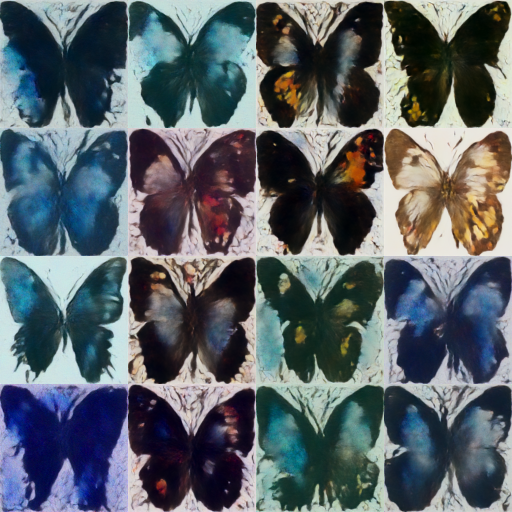}}
    \subfigure[100]{\includegraphics[width=0.11\textwidth]{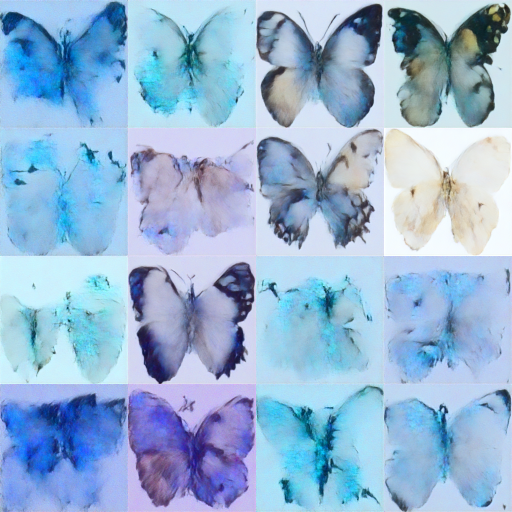}}
    \subfigure[200]{\includegraphics[width=0.11\textwidth]{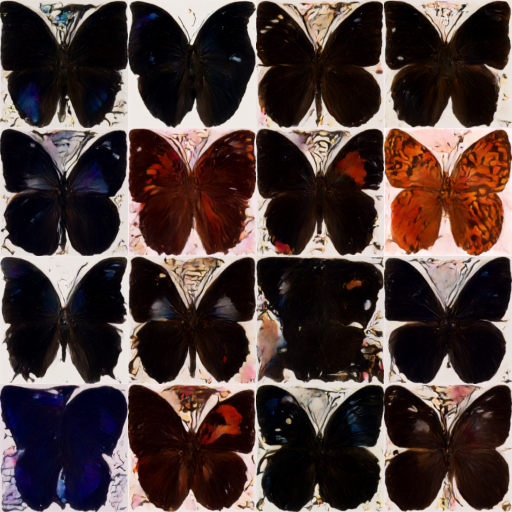}}
    \subfigure[300]{\includegraphics[width=0.11\textwidth]{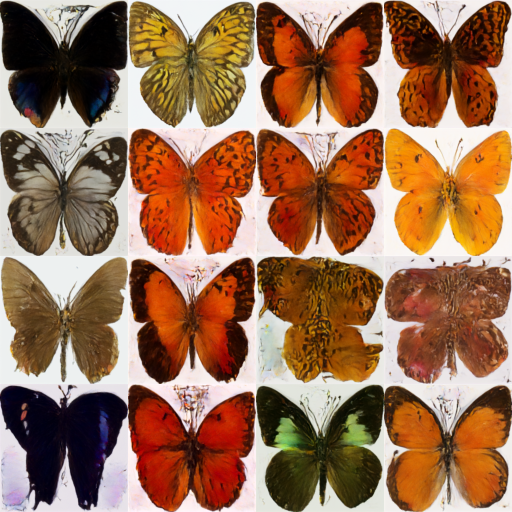}}
    \subfigure[500]{\includegraphics[width=0.11\textwidth]{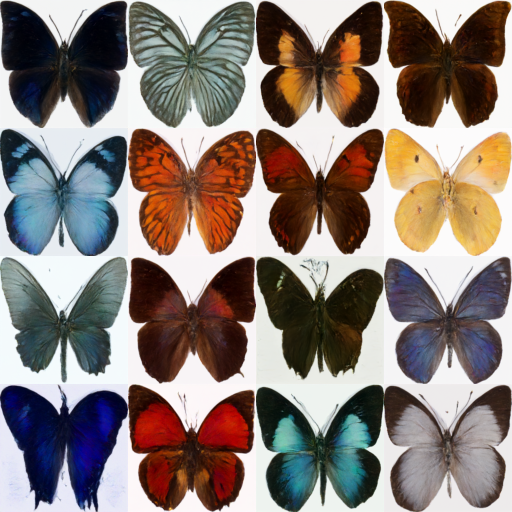}}
    \caption{Training DDPM via hyperbolic SGD (lr=2e-3) based on T sampled in a unit hyperbola, increasing epochs (10, 20, ..., 500).}
    \label{fig:ddpm_hyper_sgd_Tunit_hyperbola0.002}
\end{figure}

The image generations by Experiment SGD+LinearT, HyperSGD+LinearT, and HyperSGD+HyperT are shown in Fig. \ref{fig:ddpm_sgd_Tlinear0.002} $\sim$ \ref{fig:ddpm_hyper_sgd_Tunit_hyperbola0.002}, respectively. HyperSGD+HyperT achieves better generation butterflies at a lower training epoch than the other two cases. Moreover, HyperSGD+LinearT converges sooner than SGD+LinearT, although the generated image quality does not improve significantly.

\begin{figure}[h!]
    \centering
    \subfigure[10]{\includegraphics[width=0.11\textwidth]{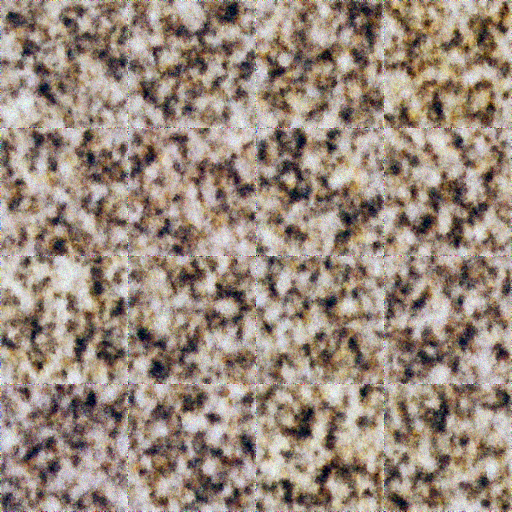}} 
    \subfigure[20]{\includegraphics[width=0.11\textwidth]{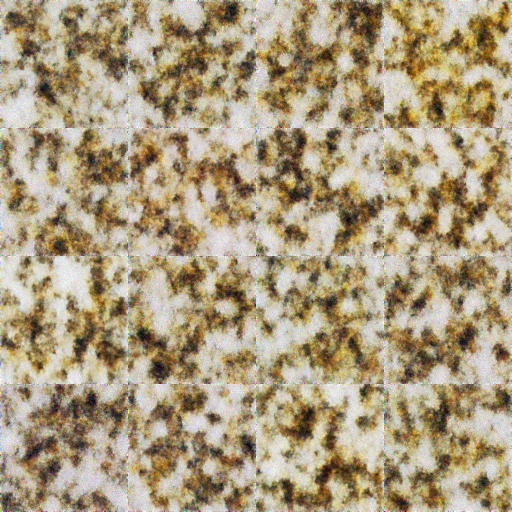}}
    \subfigure[30]{\includegraphics[width=0.11\textwidth]{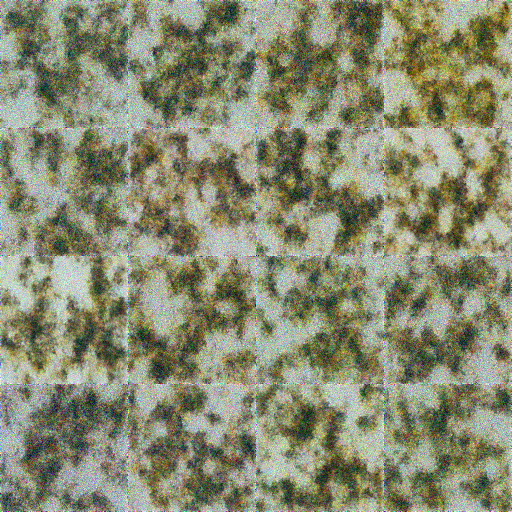}}
    \subfigure[50]{\includegraphics[width=0.11\textwidth]{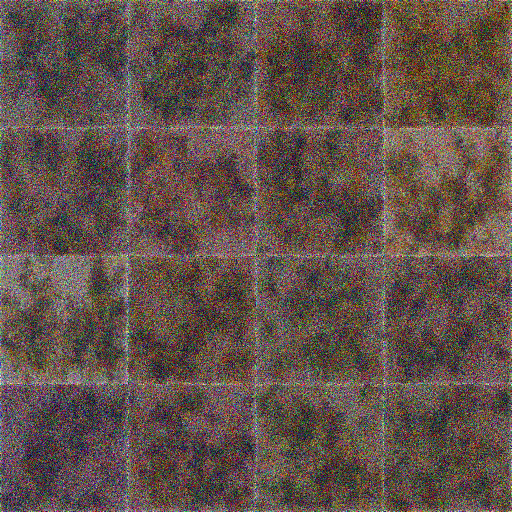}}
    \subfigure[100]{\includegraphics[width=0.11\textwidth]{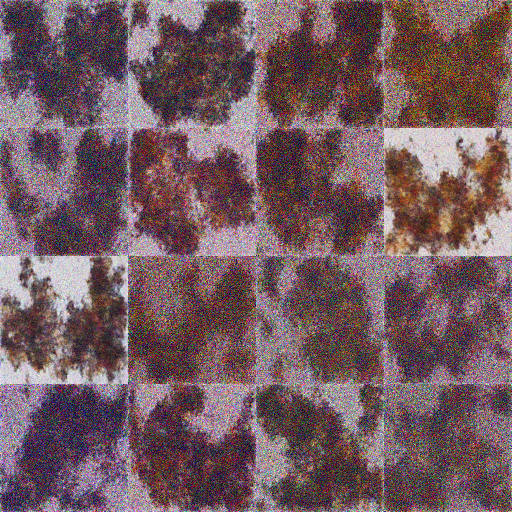}}
    \subfigure[200]{\includegraphics[width=0.11\textwidth]{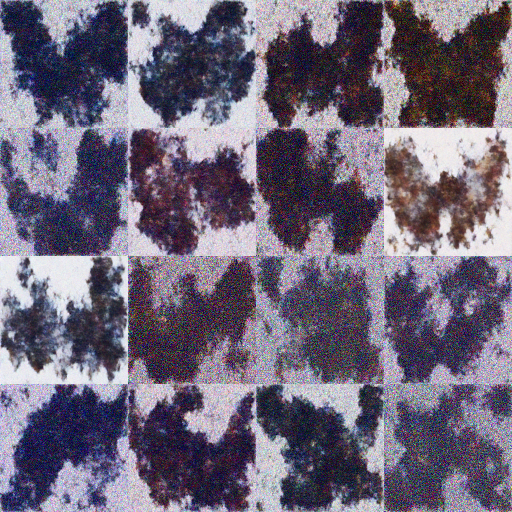}}
    \subfigure[300]{\includegraphics[width=0.11\textwidth]{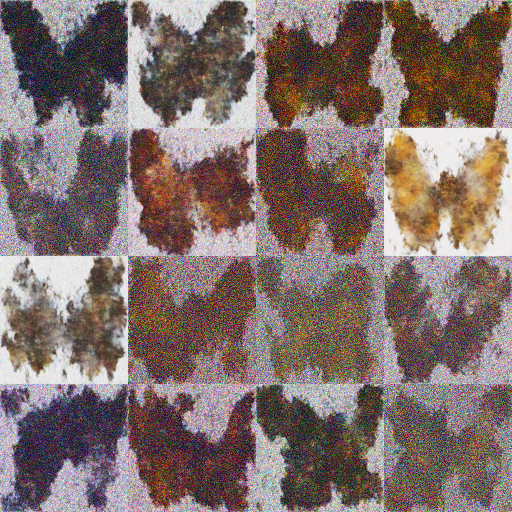}}
    \subfigure[500]{\includegraphics[width=0.11\textwidth]{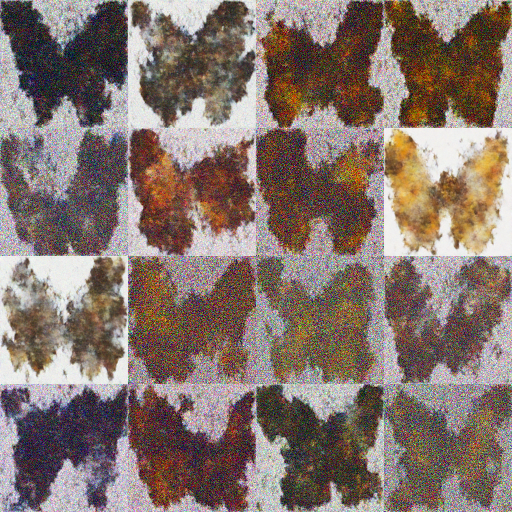}}
    \caption{Training DDPM via SGD (lr=2e-3) based on T sampled in a unit hyperbola, increasing epochs (10, 20, ..., 500).}
    \label{fig:ddpm_sgd_Tunit_hyperbola0.002}
\end{figure}

\section{Image Generation of Models Trained by AdamW}
\label{appen:gene_adamw}

\begin{figure}[h!]
    \centering
    \subfigure[10]{\includegraphics[width=0.11\textwidth]{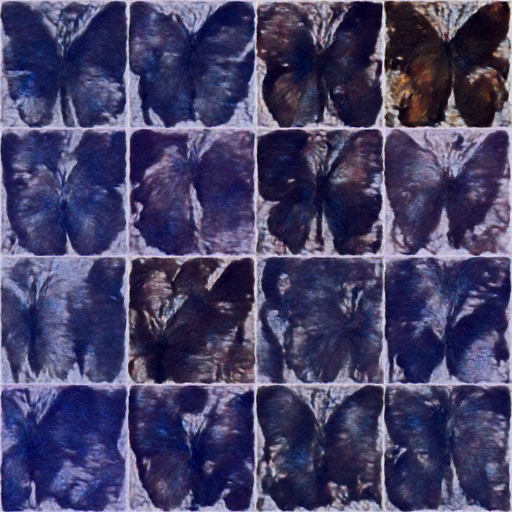}} 
    \subfigure[20]{\includegraphics[width=0.11\textwidth]{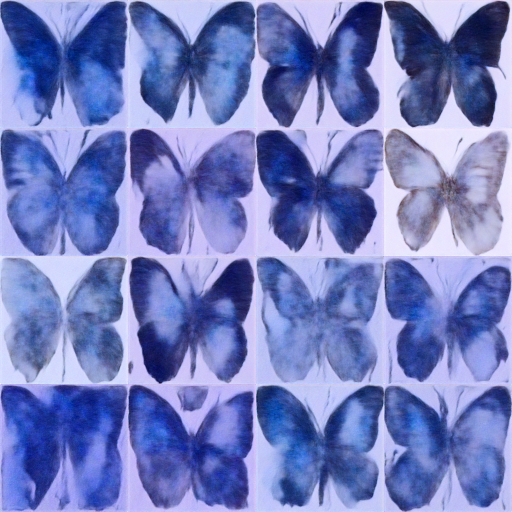}}
    \subfigure[30]{\includegraphics[width=0.11\textwidth]{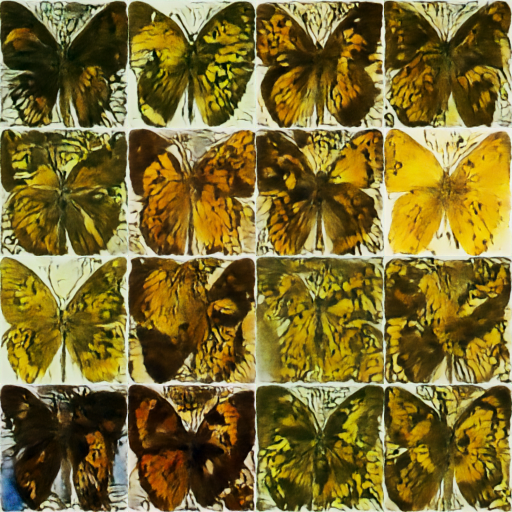}}
    \subfigure[50]{\includegraphics[width=0.11\textwidth]{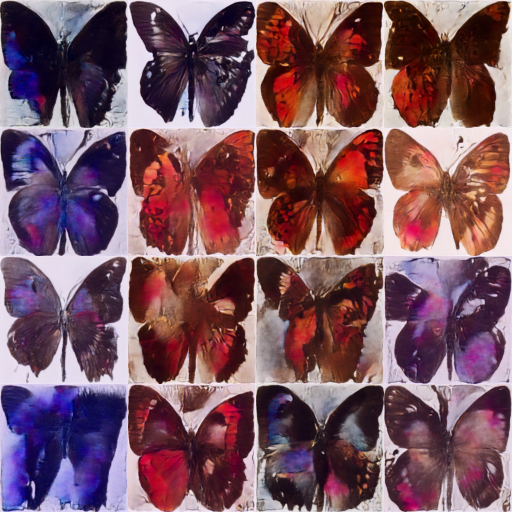}}
    \subfigure[100]{\includegraphics[width=0.11\textwidth]{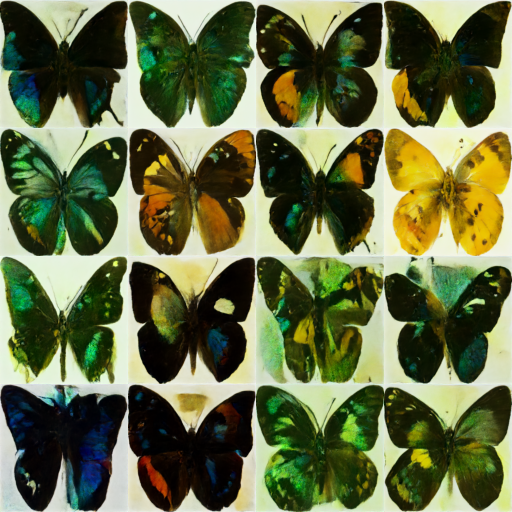}}
    \subfigure[200]{\includegraphics[width=0.11\textwidth]{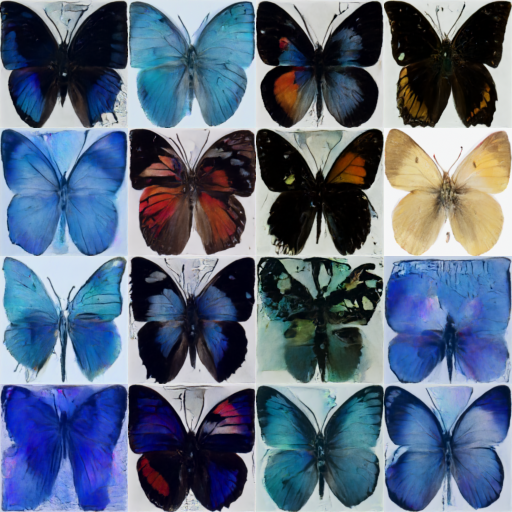}}
    \subfigure[300]{\includegraphics[width=0.11\textwidth]{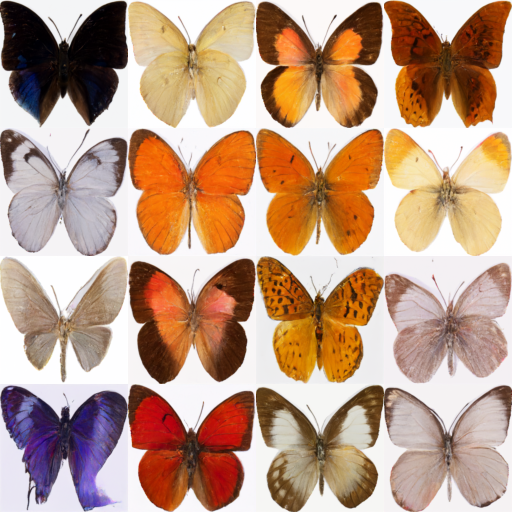}}
    \subfigure[350]{\includegraphics[width=0.11\textwidth]{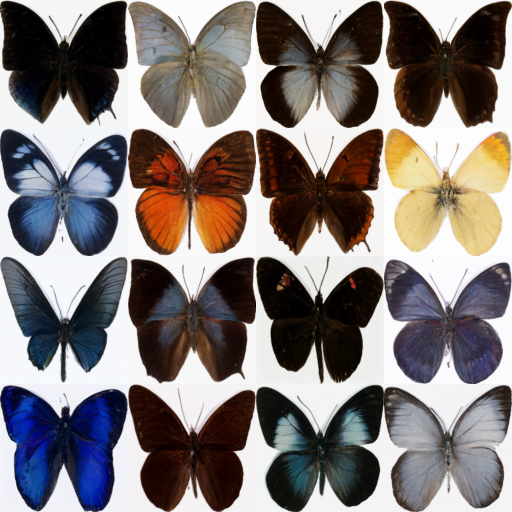}}
    \caption{Training DDPM via AdamW (lr=2e-4) based on T sampled linearly, increasing epochs (10, 20, ..., 350).}
    \label{fig:ddpm_adamw_Tlinear0.0002}
\end{figure}

\begin{figure}[h!]
    \centering
    \subfigure[10]{\includegraphics[width=0.11\textwidth]{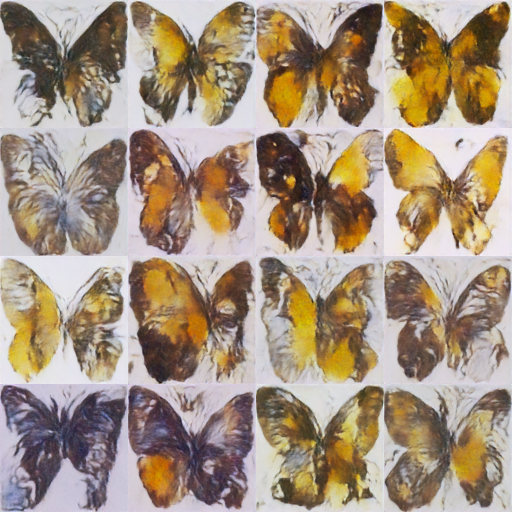}} 
    \subfigure[20]{\includegraphics[width=0.11\textwidth]{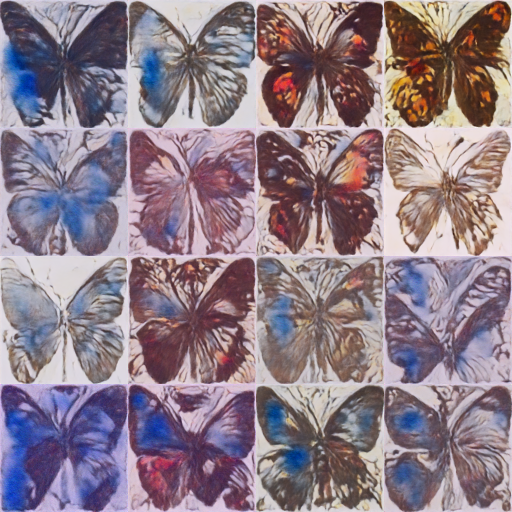}}
    \subfigure[30]{\includegraphics[width=0.11\textwidth]{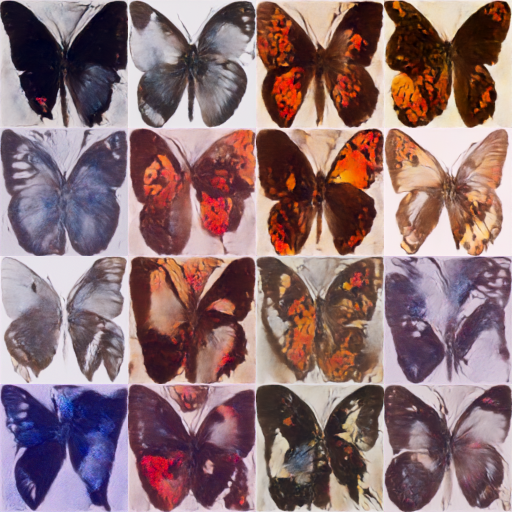}}
    \subfigure[50]{\includegraphics[width=0.11\textwidth]{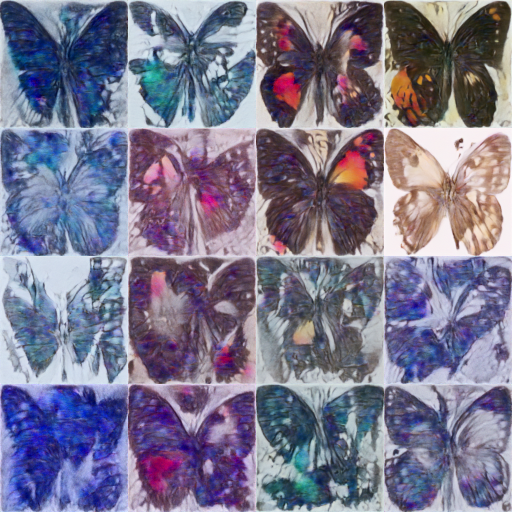}}
    \subfigure[100]{\includegraphics[width=0.11\textwidth]{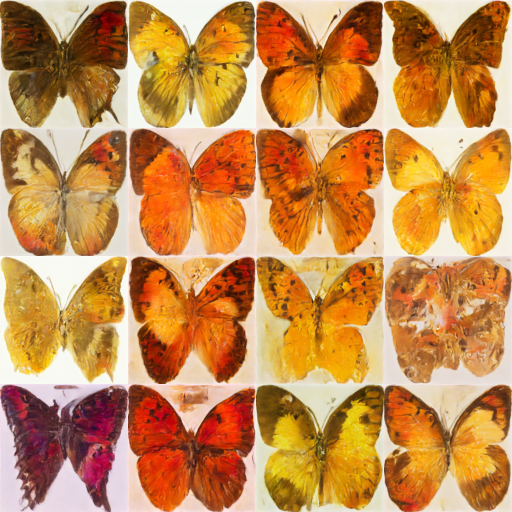}}
    \subfigure[200]{\includegraphics[width=0.11\textwidth]{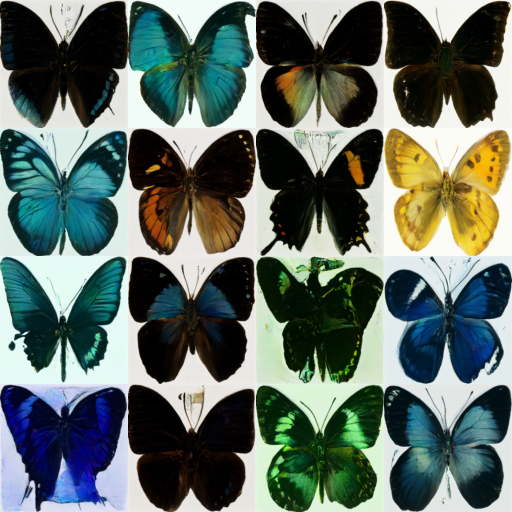}}
    \subfigure[300]{\includegraphics[width=0.11\textwidth]{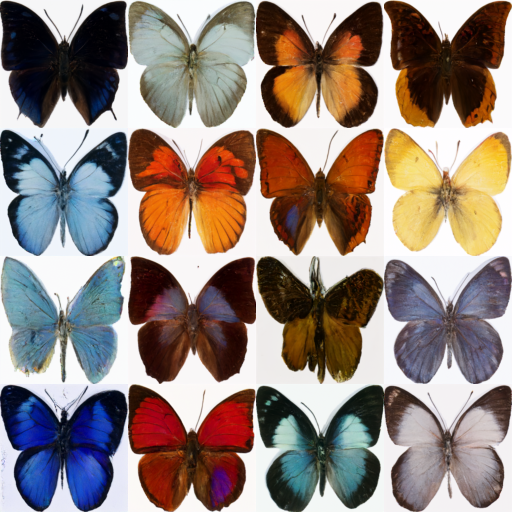}}
    \subfigure[350]{\includegraphics[width=0.11\textwidth]{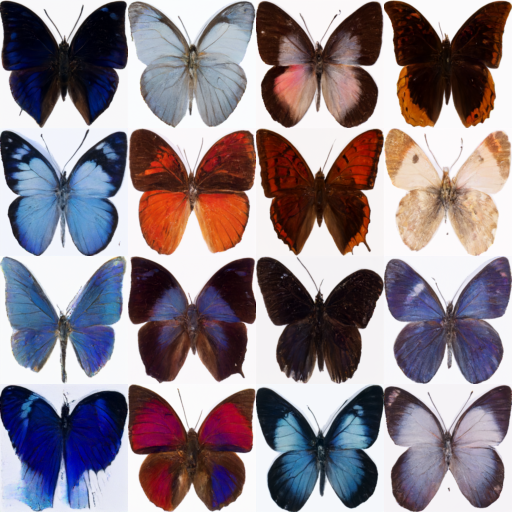}}
    \caption{Training DDPM via hyperbolic AdamW (lr=2e-4) based on T sampled in a unit hyperbola and the loss on Poincaré, increasing epochs (10, 20, ..., 350).}
    \label{fig:ddpm_hyper_adamw_Tuni_hyperbola_hyper_loss0.0002}
\end{figure}

The image generations by Experiment AdamW+LinearT, HyperAdamW+HyperT, and HyperAdamW+HyperT+HyperLoss are shown in Fig. \ref{fig:ddpm_adamw_Tlinear0.0002} $\sim$ \ref{fig:ddpm_hyper_adamw_Tuni_hyperbola_hyper_loss0.0002}, respectively. According the image quality via a visual inspection, HyperAdamW+HyperT and HyperAdamW+HyperT+HyperLoss converge slightly faster than AdamW+LinearT, as they explicit clearer butterfly shapes from the very beginning training phase.

\begin{figure}[h!]
    \centering
    \subfigure[10]{\includegraphics[width=0.11\textwidth]{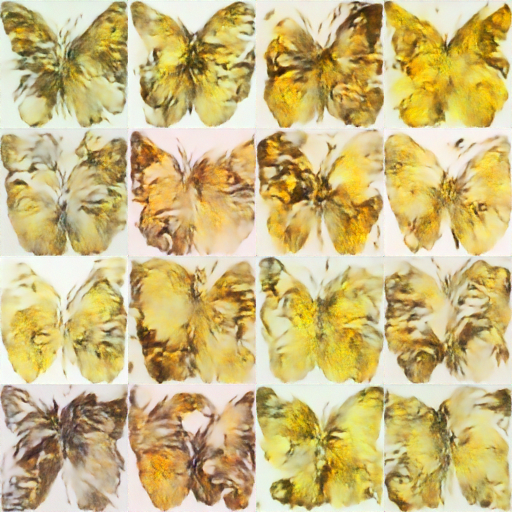}} 
    \subfigure[20]{\includegraphics[width=0.11\textwidth]{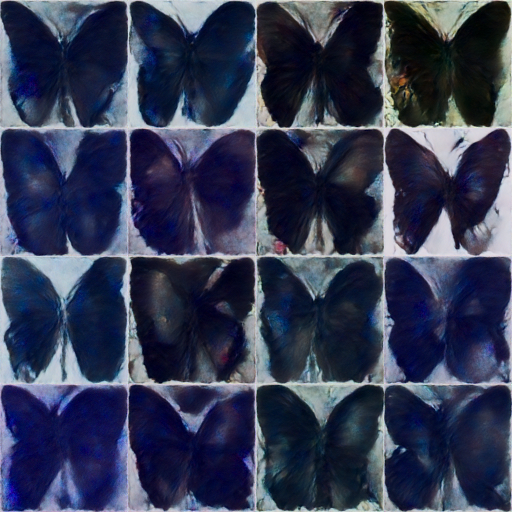}}
    \subfigure[30]{\includegraphics[width=0.11\textwidth]{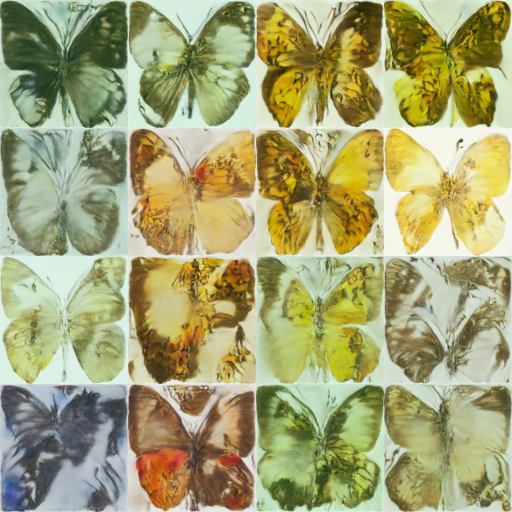}}
    \subfigure[50]{\includegraphics[width=0.11\textwidth]{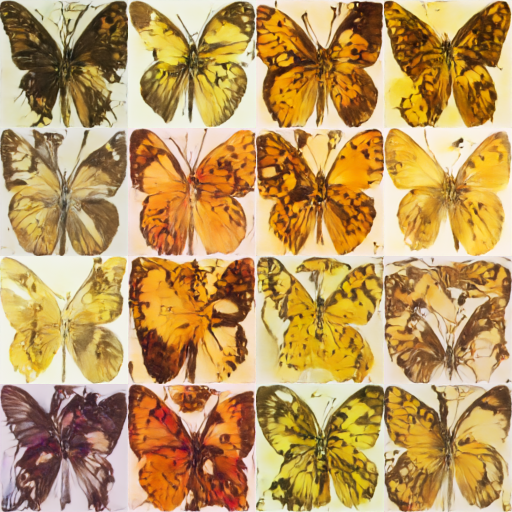}}
    \subfigure[100]{\includegraphics[width=0.11\textwidth]{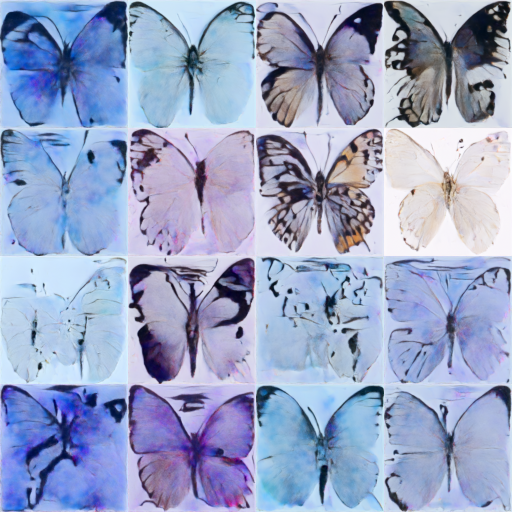}}
    \subfigure[200]{\includegraphics[width=0.11\textwidth]{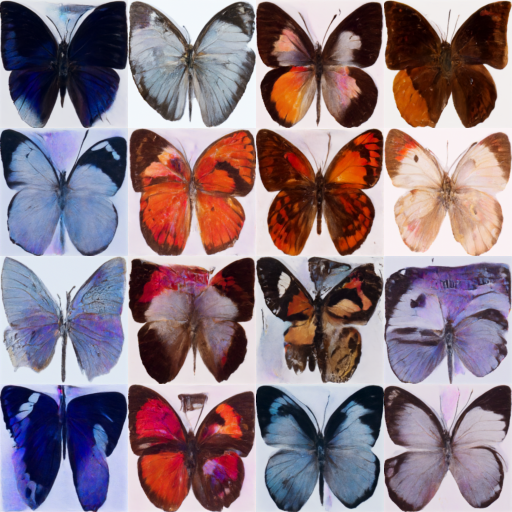}}
    \subfigure[300]{\includegraphics[width=0.11\textwidth]{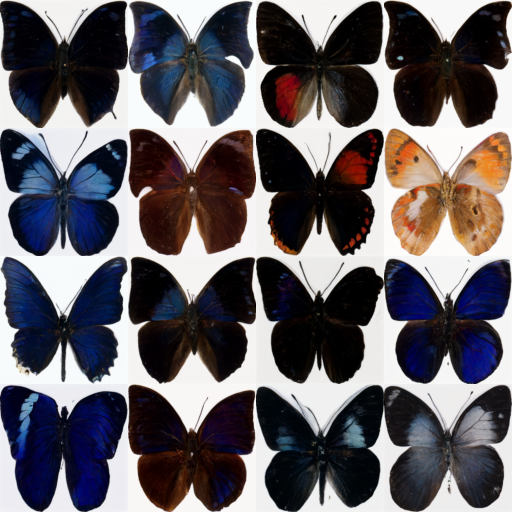}}
    \subfigure[350]{\includegraphics[width=0.11\textwidth]{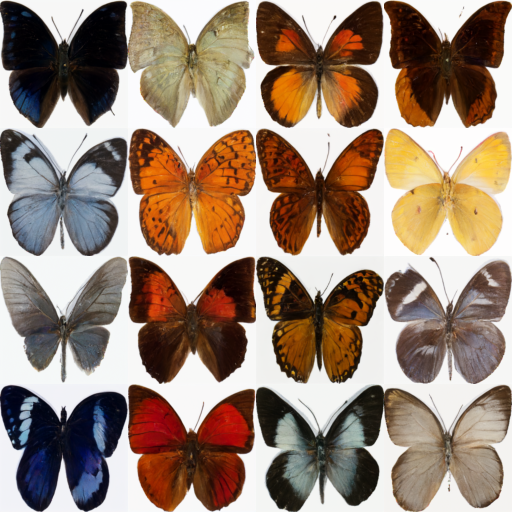}}
    \caption{Training DDPM via hyperbolic AdamW (lr=2e-4) based on T sampled in a unit hyperbola, increasing epochs (10, 20, ..., 350).}
    \label{fig:ddpm_hyper_adamw_Tuni_hyperbola0.0002}
\end{figure}

\section{Implementation and Source Codes}
\label{appen:code}

\begin{code}
\captionof{listing}{Linear and unit hyperbolic T samplers as Eq. \ref{equ:t_linear} and \ref{equ:t_unit_hyperbola}.}
\label{code:t_sampler}
\centering
\begin{minted}[mathescape, breaklines,frame=single, fontsize=\footnotesize]{python}
if t_sampler == "linear":
    self.timesteps = torch.from_numpy(np.arange(0, num_train_timesteps)[::-1].copy())
elif t_sampler == "unit_hyperbola":
    step_start = 1
    step_end = np.sqrt(num_train_timesteps**2 + 1).astype(np.float64) 
    step_sequence = np.linspace(step_start, step_end, num_train_timesteps, endpoint=True, dtype=np.float64)
    timesteps = np.sqrt(step_sequence**2 - 1).astype(np.float32)
    self.timesteps = torch.from_numpy(timesteps)
else:
    raise NotImplementedError
\end{minted}
\end{code}

\begin{code}
\captionof{listing}{Loss function on Poincaré as Eq. \ref{equ:loss_poincare}.}
\label{code:loss_poincare}
\centering
\begin{minted}[mathescape, breaklines,frame=single, fontsize=\footnotesize]{python}
def poincare_distance(pred, target, epsilon = 1e-5):
    Bs = pred.shape[0]
    loss = 0
    for i in range(Bs):
        pred_norm = mse(pred[i, :, :, :], torch.zeros_like(pred[i, :, :, :]))
        target_norm = mse(target[i, :, :, :], torch.zeros_like(target[i, :, :, :]))
        mse_loss = mse(pred[i, :, :, :], target[i, :, :, :])
        torch.arccosh(1 + 2 * mse_loss / ((1-pred_norm**2)*(1-target_norm**2) + epsilon))
        loss += mse_loss
    loss = loss / Bs
    # print(loss)
    return loss
\end{minted}
\end{code}

\begin{code}
\captionof{listing}{Transfer Euclidean gradient to hyperbolic space as Eq. \ref{equ:hyper_gt}.}
\label{code:hyper_gt}
\centering
\begin{minted}[mathescape, breaklines,frame=single, fontsize=\footnotesize]{python}
def _to_poincare(param, grad):
    return 1/4 * (1-torch.norm(param)**2)**2 * grad

def to_poincare(params, grads):
    if isinstance(params, torch.Tensor):
        return _to_poincare(params, grads)
    elif isinstance(params, list):
        num = len(params)
        for i in range(num):
            grads[i] = _to_poincare(params[i], grads[i])
        return grads
\end{minted}
\end{code}

\begin{code}
\captionof{listing}{Projection as Eq. \ref{equ:projection}.}
\label{code:projection}
\centering
\begin{minted}[mathescape, breaklines,frame=single, fontsize=\footnotesize]{python}
def proj(params, epsilon=1e-5): 
    if isinstance(params, torch.Tensor):
        return _proj(params)
    elif isinstance(params, list):
        num = len(params)
        for i in range(num):
            params[i] = _proj(params[i])
        return params
        
def _proj(param, epsilon=1e-5): 
    param_norm = torch.norm(param)
    if param_norm >=1:
        return param / param - epsilon
    else:
        return param
\end{minted}
\end{code}

\end{document}